\documentclass[runningheads]{llncs}

\usepackage{hyperref}
\usepackage[T1]{fontenc}
\usepackage{graphicx, amsmath, multirow, booktabs,appendix}
\usepackage{amssymb}
\usepackage{xcolor}
\usepackage{algorithm}
\usepackage{algpseudocode}
\usepackage{pifont}
\usepackage{dblfloatfix}
\usepackage{adjustbox}
\usepackage{cleveref}
\usepackage{orcidlink}

\usepackage{graphicx}
\begin{document}

\title{Tailored Transformation Invariance for Industrial Anomaly Detection}
\author{Mariette Sch\"onfeld\inst{1,2}\orcidlink{0009-0000-3335-6538}\and 
Wannes Meert\inst{1,2}\orcidlink{0000-0001-9560-3872} \and
Hendrik Blockeel\inst{1,2}\orcidlink{0000-0003-0378-3699}\\ \email{\{mariette.schoenfeld,wannes.meert,hendrik.blockeel\}@kuleuven.be}}
\authorrunning{M. Sch\"onfeld et al.}
\institute{
KU Leuven, Dept. of Computer Science, B-3000 Leuven, Belgium \and 
Leuven.AI - KU Leuven Institute for AI, B-3000 Leuven, Belgium}

% \subtitle{Corresponding authors: \email{mariette.schoenfeld@kuleuven.be}}

%
\maketitle              % typeset the header of the contribution
\begin{abstract}
Industrial Anomaly Detection (IAD) is a subproblem within Computer Vision Anomaly Detection that has been receiving increasing amounts of attention due to its applicability to real-life scenarios. Recent research has focused on how to extract the most informative features, contrasting older kNN-based methods that use only pretrained features. These recent methods are much more expensive to train however and could complicate real-life application. Careful study of related work with regards to transformation invariance leads to the idea that popular benchmarks require robustness to only minor translations. With this idea we then formulate LWinNN, a local window based approach that creates a middle ground between kNN based methods that have either complete or no translation invariance. Our experiments demonstrate that this small change increases accuracy considerably, while simultaneously decreasing both train and test time. This teaches us two things: first, the gap between kNN-based approaches and more complex state-of-the-art methodology can still be narrowed by effective usage of the limited data available. Second, our assumption of requiring only limited translation invariance highlights potential areas of interest for future work and the need for more spatially diverse benchmarks, for which our method can hopefully serve as a new baseline. Our code can be found at \\\href{https://github.com/marietteschonfeld/LWinNN}{https://github.com/marietteschonfeld/LWinNN}.

\end{abstract}

\section{Introduction}
\label{section:intro}

Industrial Anomaly Detection (IAD) is a subfield of computer vision that concerns the visual inspection of objects with the goal of detecting abnormalities in said object. Two specific tasks are often distinguished: predicting whether an image shows an anomaly (detection), and predicting where the anomaly is (localization). The first task is a special case of image classification, the second is more closely related to segmentation. 

The goal of practical application has led to IAD having a distinctive setting with some notable aspects. First, the training dataset contains only a few hundred normal samples: the potential variety in defects is oftentimes too large to be encapsulated in a single, unbalanced dataset without supervision. Second, an application scenario might only feature one type of object, and so a model is trained on only one class of images. Third, the anomaly detectors are often meant to be used under demanding operational settings (e.g., at least $n$ objects per second must be evaluated). Finally, the desire to segment anomalies in addition to detecting their presence has led to the realization that treating the image as a collection of patches is more effective than processing the image in its entirety. Although this setting is highly specific, the potential for real-life applications has led to a multitude of research for this problem. 

% IAD has some characteristics that distinguish it from most other computer vision tasks. First, training data is typically limited: in practical use cases, one may have only a few hundred training images, or even less. Second, it is often a one-class problem: the training data contain strictly normal cases. Third, anomaly detectors are often trained for one specific type of object, which means there is limited variability in the training data. Fourth, the anomaly detectors are often meant to be used under demanding operational settings (e.g., at least $n$ objects per second must be evaluated). Because of this specific setting, specialized methods have been developed for this area. 

One type of methods is based on the nearest neighbors principle. The reasoning is that a patch of pixels in an image is considered anomalous when it is unlike any patches seen in the training images. Using a distance metric or dissimilarity measure between feature-based representations of these patches allows for the evaluation of a patch's anomalousness. The higher the dissimilarity between a test patch and the most similar training patch observed, the more anomalous the patch. Methods can vary in their way of deriving the patch's feature representation, or which train patches the test patch is compared with. For instance, SPADE compares test patches to train patches in the same location of the image \cite{cohen2020sub}, whereas PatchCore does not limit the position of the patches compared to \cite{roth2022towards}. These methods also showed how using a pretrained feature extractor to derive feature representations for patches can result in highly accurate methodology. Moreover, the limited training data makes these nearest neighbors-based algorithms very efficient at both training and testing, dismantling the typical criticisms of nearest-neighbors (NN). More recently, deep learning-based methods that learn specialized features out of pretrained ones have been proposed, contrasting the aforementioned methods which use only pretrained features. These methods boast extremely high accuracy with low latency \cite{batzner2024efficientad,yao2025glad}, and they are considered the state-of-the-art in IAD. 

Although these deep learning-based methods outperform their NN-based predecessors at eventual deployment, both can be of merit in their own ways. Implementing an AI-based system in an industrial setting can require many iterations on aspects other than the actual ML algorithm, like data collection or integration with other parts of the system. Furthermore, an exploratory phase that shows that the benefits will outweigh the costs might be needed to justify the investment in expensive hardware. Finally, frequent personnel changes, a high system failure rate, or lack of in-house knowledge can make a model with simple maintenance more profitable than a complex one. In short, a more `heuristic' algorithm (i.e. higher interpretability, based on simple concepts, quick to train, etc.) can be of great value in industrial settings \cite{diaz2023joint}, either as a development tool or a practical compromise. 

But such heuristics can also help academic researchers sharpen their knowledge. A failing heuristic can show which areas need to be targeted with more complex methodology, while a succeeding heuristic can show specific limitations of the datasets they are tested on \cite{lipton2019troubling}. In conclusion, studying and improving the aforementioned nearest-neighbors methods can have value in many ways.

When studying these nearest-neighbors based methods carefully we noticed that transformation invariance (i.e. affine transformations like rotation) improves results on benchmarks considerably. Analysis of these benchmarks and the theory surrounding transformation invariance leads to the idea that the same gains can also be acquired by creating invariance to only minor translations. This is a much simpler problem than full transformation invariance and can be solved with relatively small changes to existing methodology. With results on benchmark datasets we demonstrate that this tailored transformation invariance makes for a more accurate and faster algorithm, narrowing the gap with more complex methodology. Finally, we analyze what we can learn from the cases where our heuristic either fails or succeeds.
% and what this means for the broader perspective on IAD. 

% As our method mainly hinges on removing elements from other works (or adding only the minimum), we end by discussing the implications for the IAD field as a whole.

% a need for more spatially diverse benchmarks. as our heuristic approach shows that transformation invariant methodology 

% We therefore propose LWinNN (Local Window Nearest Neighbor), a method that is motivated by observations regarding \textit{transformation invariance} to spatial variations in IAD datasets. These observations are made by carefully studying related work and their approach to these variations and lead us to a tentative assumption: in order to achieve good results on current popular IAD datasets with a nearest neighbor-based approach, only invariance to small translations has to be created. LWinNN does this by comparing a test patch to train patches at nearby (but not necessarily the same) locations in the training set. It thus occupies a middle ground between SPADE \cite{cohen2020sub} and PatchCore \cite{roth2022towards}, that create either no or full translation invariance. An empirical evaluation shows that this very simple change brings the performance of nearest neighbor-based methods closer to that of the deep learning-based state-of-the-art (at least for the benchmarks considered here), while remaining computationally much simpler. More importantly, our heuristic approach exposes potential areas of interest to target with future work.

\section{Why only minor translation invariance?}\label{section:preliminiaries}
\begin{figure*}[b!]
    \centering
    \caption{Visualization of our key observations w.r.t. transformation invariance in IAD.}
    \label{fig:key_observations}
    \includegraphics[width=\textwidth]{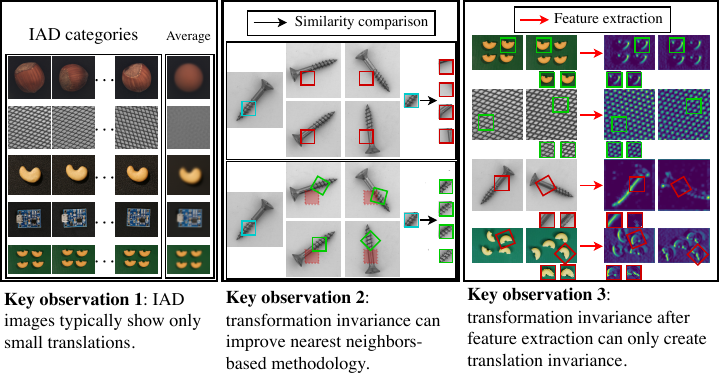}
\end{figure*}

In this section we explain why invariance to only minor translations should be sufficient in many cases. This is a logical conclusion of several observations we make w.r.t. related work, a visual demonstration of which can be found in \cref{fig:key_observations}.

\subsection{Problem statement}\label{section:problem_setting}
The goal in IAD is to find structural defects in images of objects from an industrial setting, using only a small number of normal images for training. More formally, the problem setting is as follows: given a set of images of objects, construct a function that produces for a given image an output consisting of (1) an anomaly score for each pixel in the image, and (2) an anomaly score for the image as a whole. A pixel's anomaly score should be high when that pixel deviates in a meaningful way from what is expected (i.e., it is likely part of a defect). The overall anomaly score should be high when the image is likely to show a defective object. In this paper, the term anomaly detection refers to the task of obtaining an accurate image score, and anomaly segmentation to the task of obtaining accurate pixel-wise scores. 
\begin{figure}[b!]
    \centering
    \includegraphics[width=\linewidth]{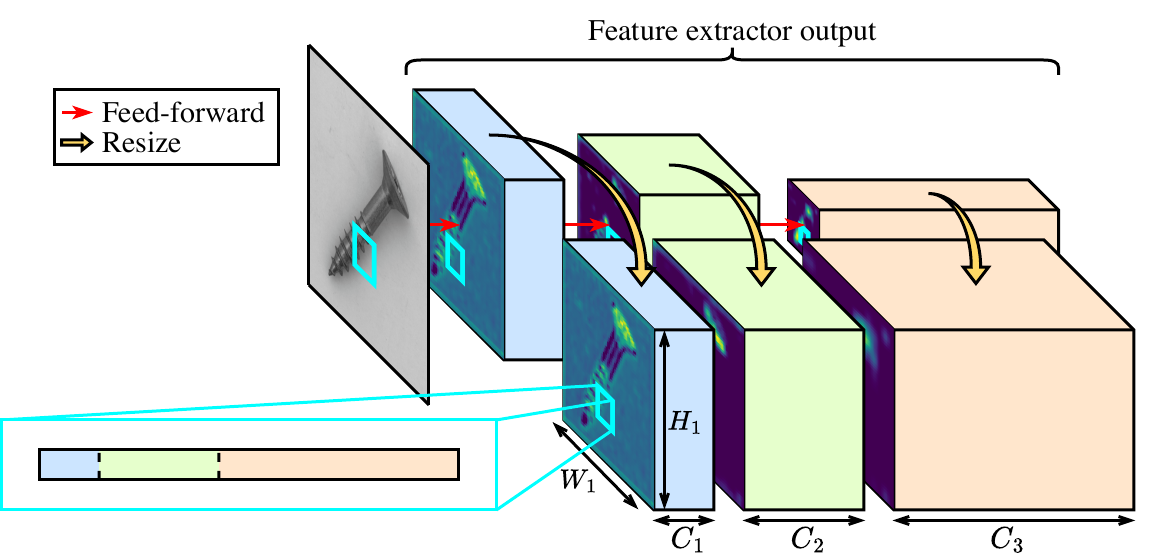}
    \caption{Example embedding generation \cite{cohen2020sub}. Resizing and concatenating the feature maps makes for straightforward access to multi-layer information for one patch of pixels.}
    \label{fig:embedding_extraction}
\end{figure}
The industrial setting of IAD implies that some assumptions can be made safely. For example, the objects to be inspected all belong to the same category, or a limited number of such categories. Objects are presented in a systematic way: they are fairly centered, placed on a non-descript background, and take up majority of the frame. Objects can be aligned differently across samples, for example any rotation of the object can constitute normal presentation. The setting also implies that translations are typically minor though: the object cannot go far before moving out of frame. A notable exception to this assumption is the multiple-instance case where multiple (smaller) objects are shown in one image, a minority of images in current IAD benchmarks. This leads us to \textbf{key observation 1: Most of the images in current IAD benchmarks show only minor translations}.

\subsection{Strides in IAD methodology}\label{section:iad_strides}
Training a neural network to extract informative features is difficult in IAD due to the small amount of training samples. Most IAD methods therefore rely on a pretrained feature extractor in some form, like a ResNet \cite{he2016deep}. The idea of using only pretrained features for IAD was popularized by SPADE \cite{cohen2020sub}. SPADE describes each individual patch of pixels using the activation level of features detected at that location (according to the feature maps constructed by the ResNet). This proved to be a very powerful tool in creating feature-representations for individual patches, and many variations have been proposed since. A general visualization of this idea can be found in \cref{fig:embedding_extraction}. 
After calculating the feature representations for all patches, SPADE then calculates patch anomaly scores using a patch-wise nearest neighbor approach: it compares a test patch's feature representation at location $(h,w)$ to all train patches' feature representation in the same location in the training images, and finds the most similar one. The similarity between test patches and their nearest neighbors is then used to calculate anomaly scores for all patches in the image. Resizing the map of patch anomaly scores to the shape of the original image then creates pixel anomaly scores. The overall anomaly score for the image is found similarly with image-wise neighbors instead of patch-wise neighbors. 

PaDiM uses the same feature representation that SPADE introduced and developed the idea that images are similar on a local level, but estimates a multi-variate Gaussian by calculating the mean and covariance matrix over the feature representations for every location $(h,w)$ \cite{defard2021padim}. The relationship between a patch and its location in the image can also be learned in other ways than assuming it explicitly like SPADE \cite{cohen2020sub} and PaDiM \cite{defard2021padim}. For example, Patch-SVDD learns which locations a patch typically occurs in \cite{yi2020patch} with self-supervision. CutPaste built on this approach by cutting a random patch from an image and pasting it in a different location to create synthetic anomalies \cite{li2021cutpaste}. 

PatchCore alleged that the relationship of a patch and its location is not that important and thus investigated transformation invariance \cite{roth2022towards}. It extracts feature representations in a manner similar to SPADE and PaDiM, but its nearest neighbor-based search compares the test patch to patches at any spatial location in the training data. By using a greedy clustering algorithm to store only the most descriptive train patches, PatchCore was able to achieve state-of-the-art accuracy while keeping training and testing cost low. This leads us to \textbf{key observation 2: adding transformation invariance to nearest neighbor-based approaches can improve results}. Further research focused on creating specialized features out of pretrained features. Successful attempts include transfer learning \cite{lee2022cfa}, knowledge distillation \cite{batzner2024efficientad}, synthetic anomalies \cite{zavrtanik2021draem}, or normalizing flows \cite{gudovskiy2022cflow}.

\subsection{Transformation (in)variance in IAD}
Although the general topic of transformation in- or equivariance is vast and varied, many of these techniques rely on a supervised task (e.g. classification) after feature extraction in order to learn which transformations are applicable to the data at hand \cite{cohen2016group,weiler2018learning}. Although the limited supervision in IAD would make these techniques complex to use effectively, methods like PatchCore demonstrated that alternative approaches to transformation invariance can be very effective and a variety of other works have followed suit. 

For example, GLAD uses an out-of-the-box alignment algorithm to align objects to a reference image \cite{artola2023glad}. This improves results, but could also destroy anomalies: some objects contain no variation in alignment within normal images, but skewed objects can constitute an anomaly. To combat this problem, FYD uses a spatial transformer network \cite{zheng2022focus}, an attention mechanism that learns typical alignment from training data \cite{jaderberg2015spatial} by an alignment-based loss. RegAD continued with spatial transformers \cite{huang2022registration}, but placed the alignment module \textit{after} feature extraction instead of \textit{before} like GLAD and FYD. This is noteworthy, because alignment after feature extraction can only result in translation invariance \cite{finnveden2021understanding}. Feature extraction has only guaranteed invariance to translation. Techniques like max-pooling can create minor robustness to other transformations, but feature maps of two objects that are identical apart from a rotation can look significantly different. This leads to \textbf{key observation 3: adding transformation invariance after feature extraction can only create translation invariance}. 

Although this observation may appear like we imply that post-feature extraction transformation is incorrect, we see that all methods that introduce transformation invariance (either before or after feature extraction) see a significant improvement in results. This suggests that translation invariance can be sufficient for good performance on current benchmarks.

\section{How to achieve minor translation invariance}\label{section:method}
% embedding extraction

We reiterate our three key observations from the previous section: First, the majority of translations we expect to encounter in (current) IAD benchmarks are minor. Second, transformation invariance can benefit nearest neighbor-based approaches. Third and finally, creating transformation invariance after feature extraction can only result in translation invariance. We can therefore theorize that if we want to receive the same benefits of transformation invariance that other nearest neighbor-based methods get, we might only need to create invariance to minor translations. This is a much simpler problem than full transformation invariance and can be achieved with small, but targeted changes to existing methodology.

We start by creating feature representations for all patches by forming an embedding of feature maps. We then perform a nearest neighbor search in a local window to calculate patch anomaly scores. Aggregating patch anomaly scores results in an image anomaly score. 
\subsection{Embedding generation}
We use the same method of embedding generation that PatchCore uses \cite{roth2022towards}, a refinement of techniques proposed by SPADE \cite{cohen2020sub}. In this method an image is fed through a pretrained neural network. The output of the first $i$ layers is set aside as feature maps. These feature maps are then resized and interpolated to the size of the first (and largest) feature map. Concatenating the maps on the channel dimension creates an embedding for the original image. This process is illustrated in \cref{fig:embedding_extraction}. Like PatchCore, we also use a pooling operation before resizing the feature maps. 

More formally, an image with shape $C_0 \times H_0 \times W_0$ is fed through a pretrained neural network, setting aside the output of the first $i$ layers as feature maps, which have shape $C_i\times H_i \times W_i$ for every layer $i>0$. These maps are pooled, resized to $H_1 \times W_1$ pixels, and concatenated on the channel dimension. This tensor forms the image's embedding, and has shape $C \times H_1 \times W_1$ with $C=\sum_i C_i$. Finally we stack all $N_{\text{train}}$ embeddings, creating a tensor $X^{\text{train}}$ of shape $N_{\text{train}} \times C \times H_1 \times W_1$. 
\begin{figure}[b!]
    \centering
    \includegraphics[scale=0.75]{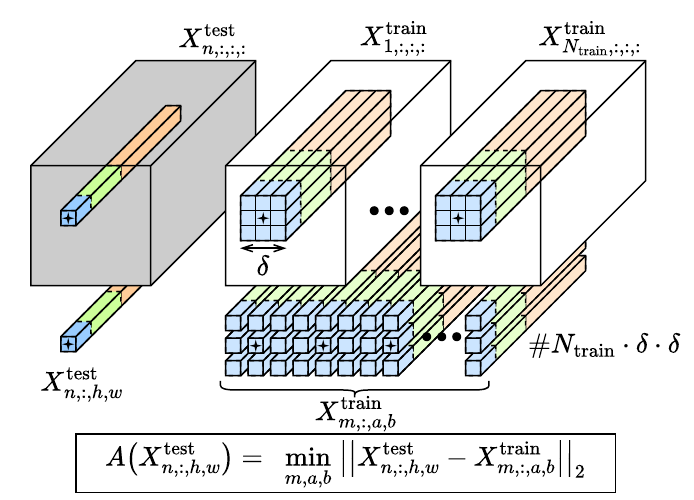}
    \caption{A visualization of local window nearest neighbor. For every test feature vector, we do a 1NN search in a local window of size $\delta\cdot \delta$.}
    \label{fig:nearest_neighbor}
\end{figure}
\subsection{Anomaly segmentation with local window nearest neighbor}
This embedding of equally sized feature maps simplifies accessing a patch's multi-layer feature representation: the representation for patch location $(h,w)$ of image $n$ corresponds to the vector $X^{\text{train}}_{n,:,h,w}$. Creating a test embedding $X^{\text{test}}$ in the same way allows for calculating the dissimilarity of two patches by comparing their corresponding feature vectors. For example, SPADE \cite{cohen2020sub} calculates patch anomaly scores with local 1-nearest neighbor:
\begin{equation}
\begin{split}
    A\left(X^{\text{test}}_{n,:,h,w}\right) = &\min_{m} \| X^{\text{test}}_{n,:,h,w} -  X^{\text{train}}_{m,:,h,w}\|_2, \\&m \in \left\lbrack 0,N_{\text{train}} \right\rbrack
\end{split}
\end{equation}
This method is not robust to translations. PatchCore targets this problem by comparing a test feature vector to a coreset of the most descriptive train feature vectors, regardless of their location \cite{roth2022towards}. This could lead to complete translation invariance, but also requires coreset generation for tractability. By searching in a local window we can avoid coreset generation and create invariance to minor translations simultaneously:
% Because we want to create invariance to only minor translations, we can avoid coreset generation by simply searching locally: 
\begin{equation}
    \begin{split}
    A\left(X^{\text{test}}_{n,:,h,w}\right) = & \min_{m,a,b} \| X^{\text{test}}_{n,:,h,w} - X^{\text{train}}_{m,:,a,b}\|_2,\\ m &\in \left\lbrack0,N_{\text{train}} \right\rbrack, \\
                     a &\in \left\lbrack h-\left\lfloor\frac{\delta}{2}\right\rfloor,  
                     h+ \left\lceil\frac{\delta}{2}\right\rceil \right\rbrack, \\
                      b &\in \left\lbrack w-\left\lfloor\frac{\delta}{2}\right\rfloor,w+ \left\lceil\frac{\delta}{2}\right\rceil \right\rbrack              
\end{split}
\end{equation}
% local window nearest neighbor
With $\delta$ the size of this local window. Local window nearest neighbor is illustrated in \cref{fig:nearest_neighbor}.

\subsection{Anomaly detection}
Local window nearest neighbors provides patch-wise anomaly scores, but we need to output both patch-anomaly scores and an overall image anomaly score. For example, SPADE finds the $K$ most similar train images for a test image and calculates the average Euclidean distance between their embeddings \cite{cohen2020sub}:
\begin{equation}\label{eq:spade_detection}
    A\left(X^{\text{test}}_{n,:,:,:}\right) = \frac{1}{K}\sum_{k}\| X^{\text{test}}_{n,:,:,:} - X^{\text{train}}_{k,:,:,:}\|_2
\end{equation}
PaDiM argued that this method is suboptimal, as high patch anomaly scores are more indicative of the presence of anomalies, and suggests taking the maximum patch anomaly score \cite{defard2021padim}:
\begin{equation}\label{eq:lwinnn_detection}
    A\left(X^{\text{test}}_{n,:,:,:}\right) = \max_{h,w} A\left(X^{\text{test}}_{n,:,h,w}\right)
    % ,h \in \left \lbrack 0,H-1 \right \rbrack, w \in \left \lbrack 0,W-1 \right \rbrack)
\end{equation}
Though the maximum value of the patch anomaly scores may be reductive in some cases (e.g. multiple anomalous areas, or the size of an area), it  is a popular technique because it tends to provide good empirical results \cite{batzner2024efficientad,huang2022registration,roth2022towards}. As such, we use this technique as well. 

\section{Experiments}\label{section:experiments}
In this section we test our method, nicknamed LWinNN, on two popular IAD benchmarks. We compare our performance with both similar methodology and state-of-the-art algorithms before performing an ablation study. Finally we conclude with a discussion on the implications of our heuristic. 

\subsection{Details}
\paragraph{\textbf{Metrics}}
We measure accuracy with metrics standard in IAD. Anomaly detection is measured with AUROC, the Area Under the ROC-curve. For anomaly segmentation we use the IAD-specific AUPRO-score (Area Under the Per-Region Overlap curve) \cite{bergmann2019mvtec}.
% AUROC is considered to be an unsuitable metric for anomaly segmentation \cite{bergmann2019mvtec, defard2021padim}, as it is insensitive to small anomalous areas. 
% . The AUPRO-score assesses the overlap of the true anomalous areas with the predicted anomalous areas, and weighs these areas equally regardless of size.\\
We measure testing time and the length of an entire train-test cycle in seconds. Testing time is measured in seconds per image.
% dataset info
\begin{table}[b]
\centering
\caption{Details on the benchmark datasets. The number of samples is the average over all categories per type.}
\label{table:datasets}
\begin{adjustbox}{max width=\linewidth}
\begin{tabular}{llll}
\toprule
\textbf{Benchmark} & \textbf{Types (\#categories)} & \textbf{\#Train} & \textbf{\#Test (+/-)}\\
\toprule
\multirow{2}{*}{MVTec-AD \cite{bergmann2019mvtec}}   & Textures (5) & 253.2 & (26.6/76.4) \\
 & Objects (10) & 236.3 & (33.4/87.6 )\\
\midrule                  
\multirow{3}{*}{VisA \cite{zou2022spot}} & Complex structure (4) & 903.5 & (100.5/100)\\
 & Single-instance (4)& 450.75 & (90/100)\\
 & Multi-instance (4) & 810.5 & (50/100)\\
\bottomrule
\end{tabular}
\end{adjustbox}
\end{table}
\paragraph{\textbf{Benchmarks}}
The two benchmarks we use are MVTec-AD \cite{bergmann2019mvtec} and VisA \cite{zou2022spot}. Both datasets provide a train-test split. For the sake of memory consumption, we only use the first 750 images for categories that have more samples than this.
\Cref{table:datasets} details further specifics about these benchmarks. 

\paragraph{\textbf{Implementation details}}
For feature extraction we use the first three layers of a ResNet18 \cite{he2016deep} that has been pretrained on ImageNet \cite{deng2009imagenet}, implemented in the Torchvision library \cite{marcel2010torchvision}. ResNet18 is not as commonly used in IAD as Wide-ResNet50 or Wide-ResNet101, which typically provide higher accuracy. ResNet18 has significantly smaller dimensions though, which greatly reduces both train and test time.

For preprocessing we resize the shortest side of the image to 256 pixels, normalize with a scale-shift normalization, and interpolate with bilinear interpolation when resizing the feature maps. These choices are motivated in the ablation study in \cref{section:ablation_preprocessing}. We also use common IAD postprocessing steps on the patch anomaly scores to create proper anomaly maps \cite{cohen2020sub,batzner2024efficientad,roth2022towards}. These steps include resizing the maps to the size of the original image with bilinear interpolation before smoothing them using a Gaussian blur with $\sigma=4$. 

LWinNN can be implemented with an unfolding operation on every window. This enables calculation of Euclidean distances with a batched matrix multiplication over all windows simultaneously. Though efficient, this method is brittle: it easily causes GPU memory overflow. We therefore pad our train embeddings, iterate over the shape of our window, and compare the test embedding with a different slice of the train embeddings in every iteration. Our hardware setup is an NVIDIA RTX A5000 GPU with a 256 GB RAM CPU. 

\subsection{Comparison to state-of-the-art}
\paragraph{\textbf{Details}}
We compare LWinNN with window size $\delta=7$ to four other algorithms. The workings of the first three, SPADE \cite{cohen2020sub}, PaDiM \cite{defard2021padim}, and PatchCore \cite{roth2022towards} have already been discussed in detail in \cref{section:iad_strides} and \cref{section:method}. We make several small implementation changes to these methods: for SPADE we use \cref{eq:lwinnn_detection} for anomaly detection instead of \cref{eq:spade_detection}. For PaDiM we remove all data augmentation. For PatchCore we use a coreset sampling ratio of 0.01 rather than 0.1 in order to give a fairer comparison in terms of training time. For both PaDiM and PatchCore we remove centercropping, as this also requires cropping the ground truth segmentation masks. 

To also compare with current state-of-the-art accuracy, we choose EfficientAD \cite{batzner2024efficientad} as our fourth algorithm. EfficientAD uses a teacher-student network (more commonly known as knowledge distillation), where the student network imitates the teacher on IAD training images, but is punished when imitating beyond training data. This is achieved by training on a subset of ImageNet \cite{deng2009imagenet} parallel to IAD datasets. An autoencoder is trained on top of the teacher-student network to learn relationships between features and locations. We have split the cost of training the teacher network equally over all categories.

All of these methods are reproduced with ResNet18 as the pretrained feature extractor. We do not change layer selection. The combination of these changes explains why some of the results from our reproductions differ from the original paper. Removing data augmentation and centercropping worsen performance particularly, but they are considered to be controversial transformations as they require the assumption that these operations do not introduce or remove anomalies \cite{batzner2024efficientad}. 
% timings figure
\begin{table*}[t!]
\caption{Comparison of LWinNN ($\delta=7$) to multiple IAD methods. Results are reported as (detection/segmentation).}
\label{table:comparison}
\centering
\begin{adjustbox}{max width=\textwidth}\begin{tabular}{lllllll}
\toprule
Dataset & Category type & \textbf{SPADE} \cite{cohen2020sub} & \textbf{PaDiM} \cite{defard2021padim} & \textbf{PatchCore} \cite{roth2022towards} & \textbf{EfficientAD} \cite{batzner2024efficientad} & \textbf{LWinNN} \\
\midrule
\multirow[c]{3}{*}{MVTec-AD \cite{bergmann2019mvtec}} & Objects & (86.3/92.9) & (85.0/92.2) & (96.5/90.8) & (\textbf{98.2}/93.9) & (\textbf{98.2}/\textbf{94.6}) \\
 & Textures & (86.8/83.2) & (94.8/89.7) & (97.2/90.1) & (\textbf{99.6}/\textbf{91.9}) & (98.3/90.8) \\
\cline{2-7} & Average & (86.5/89.7) & (88.2/91.4) & (96.8/90.6) & (\textbf{98.7}/93.2) & (98.2/\textbf{93.4}) \\
\cline{1-7}
\multirow[c]{4}{*}{VisA \cite{zou2022spot}} & Complex structure & (92.9/88.4) & (84.1/84.0) & (96.1/84.6) & (\textbf{98.7}/\textbf{93.2}) & (98.0/92.5) \\
 & Multiple instances & (71.5/84.4) & (74.0/78.0) & (86.4/86.2) & (\textbf{94.8}/\textbf{95.9}) & (89.4/92.7) \\
 & Single instance & (90.2/77.6) & (89.0/83.9) & (96.4/85.9) & (98.0/88.8) & (\textbf{98.4}/\textbf{89.1}) \\
\cline{2-7} & Average & (84.9/83.5) & (82.4/82.0) & (93.0/85.6) & (\textbf{97.2}/\textbf{92.7}) & (95.3/91.4) \\
\cline{1-7}
Average &  & (85.7/86.6) & (85.3/86.7) & (94.9/88.1) & (\textbf{97.9}/\textbf{92.9}) & (96.7/92.4) \\
\cline{1-7}
\bottomrule
\end{tabular}\end{adjustbox}
\end{table*}

\begin{figure*}[b!]
    \centering
    \includegraphics[scale=0.8]{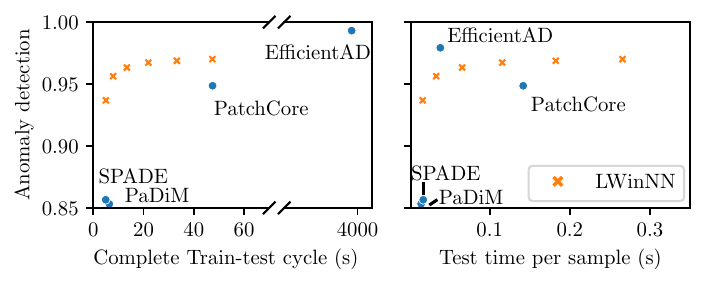}
    \caption{Train and test time plotted against anomaly detection for different IAD methods. For test time we plot LWinNN with multiple values for $\delta$. Detection is the average over both benchmarks.}
    \label{fig:scores_comparison_time}
\end{figure*}
\paragraph{\textbf{Results}}
% comparative results
The results for anomaly detection and localization in \cref{table:comparison} show that LWinNN is more accurate than the other heuristic-like methods like SPADE, PaDiM, and PatchCore. It is outperformed by EfficientAD in terms of accuracy, but the margins for most categories are relatively small. Considering the train and test time in combination with anomaly detection in \cref{fig:scores_comparison_time} shows that LWinNN provides an excellent compromise between all three metrics.
\subsection{Impact of local window-based search}
\label{subsec:local_window_search}
% window size 
\begin{figure*}[b!]
    \centering
    \includegraphics[scale=0.8]{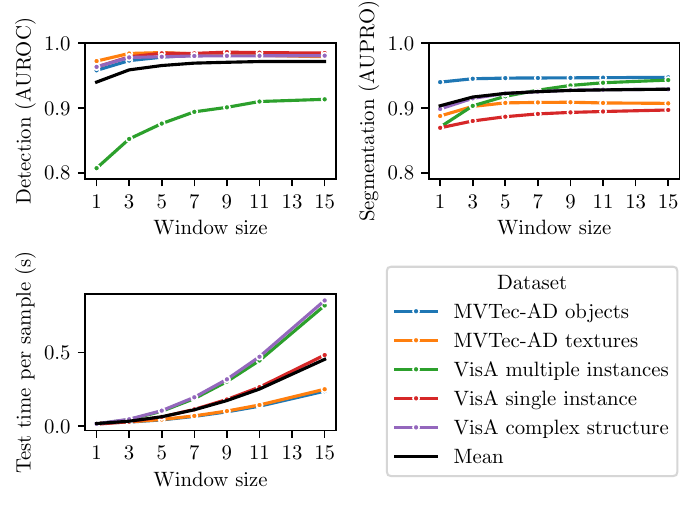}
    \caption{Performance of LWinNN over various window sizes.}
    \label{fig:scores_window_size}
\end{figure*}

We now investigate the importance of local windows by running LWinNN while varying over window size $\delta$. We plot detection scores, segmentation scores, and average test time per image in \cref{fig:scores_window_size}. Training time is not impacted by window size and we do not include it here. 
\paragraph{\textbf{Results}} 
We see that window size positively affects results, but that the effect is more pronounced for certain categories. Furthermore, the effect stagnates with larger window sizes for most categories. To put window size in perspective, an image of $256\times256$ pixels results in a $H$ and $W$ of $62\times62$, and so a window size of $\delta=5$ creates robustness to translations of only 4\% of the image height/width, in any direction. In fact, only the multiple instances categories from VisA \cite{zou2022spot} seem to benefit significantly from window sizes above 5. This reaffirms our initial observations from \cref{section:problem_setting} that only these categories feature larger translations. We also see that larger window sizes increase testing time, which is to be expected as the amount of vectors to be compared to increases quadratically with window size. The amount of training samples also negatively affects test time, like any nearest neighbors-based method. 

\subsection{Embedding generation}\label{section:ablation_preprocessing}
Many methods feature similar procedures for creating embeddings out of (pretrained) features. The broad idea of resizing and concatenating feature maps is the same, but there can be differences in minor details. Although some of these details are regularly studied with consistent results (e.g. image size, extractor and layer selection \cite{heckler2023exploring,roth2022towards,defard2021padim}), some are not and their impact on scores is unknown. This can make it difficult to study the source of gains, and so we provide a study here of these details.

\paragraph{\textbf{Details}} The differences we study are normalization, aspect ratio when resizing non-square images, pooling features after extraction, and the interpolation mode used when resizing feature maps. For every difference, we explore two choices that are typically seen in literature. For normalization we explore no normalization versus the shift-scale normalization recommended for \href{https://pytorch.org/vision/stable/models.html}{pretrained Torchvision models}. For resizing we compare resizing the image to $256 \times 256$ pixels versus preserving aspect ratio when resizing the shortest side to 256 pixels. For pooling we compare no pooling with a 1-strided 2D average pool with kernel size 3. For interpolation modes we compare between bilinear and nearest interpolation. We use a window size of 5. We test all combinations of choices, but displaying every combination in a single table complicates interpretation of results. We therefore aggregate the results over all combinations for every step, but split over the two choices. The results are shown in \cref{table:ablation}. 
\begin{table*}[t]
\caption{Ablation study for embedding generation details. Results are aggregated over all combinations and reported as (detection/segmentation)$\pm$(standard deviation/standard deviation). Results for MVTec-AD with aspect ratio preservation are omitted as all images in this dataset are square.}
\label{table:ablation}
\centering
\begin{adjustbox}{max width=\textwidth}\begin{tabular}{llll}
\toprule
 &  & MVTec-AD \cite{bergmann2019mvtec} & VisA \cite{zou2022spot} \\
\midrule
\multirow[c]{2}{*}{Interpolation mode} & Bilinear & (\textbf{94.8}/\textbf{91.9})$\pm$(6.0/5.9) & (\textbf{93.2}/\textbf{88.7})$\pm$(6.9/8.7) \\
 & Nearest & (94.5/91.5)$\pm$(6.2/6.3) & (92.4/88.2)$\pm$(7.4/8.1) \\
\cline{1-4}
\multirow[c]{2}{*}{Normalization} & \ding{55} & (94.6/90.6)$\pm$(5.6/6.5) & (92.0/87.5)$\pm$(7.3/9.1) \\
 & \ding{51} & (\textbf{94.7}/\textbf{92.8})$\pm$(6.6/5.5) & (\textbf{93.6}/\textbf{89.4})$\pm$(7.0/7.5) \\
\cline{1-4}
\multirow[c]{2}{*}{Pooling} & \ding{55} & (91.7/91.6)$\pm$(7.0/7.0) & (92.4/88.2)$\pm$(6.1/9.5) \\
 & \ding{51} & (\textbf{97.5}/\textbf{91.9})$\pm$(3.0/5.0) & (\textbf{93.2}/\textbf{88.7})$\pm$(8.1/7.1) \\
\cline{1-4}
\multirow[c]{2}{*}{Aspect ratio preservation} & \ding{55} & - & (92.4/87.7)$\pm$(7.7/8.5) \\
 & \ding{51} & (\textbf{94.6}/\textbf{91.7})$\pm$(6.1/6.1) & (\textbf{93.2}/\textbf{89.2})$\pm$(6.7/8.2) \\
\cline{1-4}
\bottomrule
\end{tabular}\end{adjustbox}
\end{table*}

\paragraph{\textbf{Results}}
We see that using normalization, preserving aspect ratio and pooling all offer a clear benefit for performance, though the gain achieved with pooling is a lot more significant than normalization and aspect ratio preservation. For interpolation mode the results point slightly in favor of bilinear interpolation, but the difference is too small to draw a definitive conclusion. That the combination of choices does matter is clear from our results though: our combination (bilinear interpolation, normalization, pooling, and aspect ratio preservation) results in an average detection score of 96.34, while the opposite of combination (nearest interpolation, no normalization, no pooling, no aspect ratio preservation) results in 91.22. 

For relatively small design decisions, these differences are noteworthy. Usage of shift-scale normalization is rarely mentioned in IAD papers, and its usage or lack thereof is only shown through supplementary material. By using it we improve our average detection score from 93.3\% to 94.2\%, a gain large enough to be considered significant. Although this is an obvious transformation for many in computer vision, researchers coming from anomaly detection might be unaware that this form of normalization is considered standard. To illustrate the usage of this transformation, we have considered a sample of 11 IAD papers \cite{defard2021padim,cohen2020sub,roth2022towards,huang2022registration,jang2023n,lee2022cfa,zavrtanik2021draem,batzner2024efficientad,yi2020patch,liu2023simplenet,gudovskiy2022cflow} and investigated whether they mention this transform in their paper or supplementary material. Of the seven that published code, four used it \cite{roth2022towards,lee2022cfa,liu2023simplenet,gudovskiy2022cflow}, one commented this transformation \cite{huang2022registration}, and two did not use it \cite{zavrtanik2021draem,yi2020patch}. Out of the four papers without code, only one mentioned using this normalization in their appendix \cite{batzner2024efficientad}. 

None of these preprocessing choices constitute an actual contribution, but the gains they demonstrate here show that a fair comparison between methods should perhaps involve standardization. Further research is needed to evaluate whether the impact of these differences is universal across other methods before suggesting such a standardization however. 

\subsection{Multi-instance cases}
\begin{table*}[b]
\caption{Comparison of multiple IAD methods on VisA multiple instances categories \cite{zou2022spot}. Results are reported as (detection/segmentation).}
\label{table:comparison_multi_instance}
\centering
\begin{adjustbox}{max width=\textwidth}\begin{tabular}{llllll}
\toprule
Category & \textbf{SPADE} \cite{cohen2020sub} & \textbf{PaDiM} \cite{defard2021padim} & \textbf{PatchCore} \cite{roth2022towards} & \textbf{EfficientAD} \cite{batzner2024efficientad} & \textbf{LWinNN}  \\
\midrule
Candle & (76.9/94.9) & (84.8/94.5) & (95.3/90.4) & (\textbf{96.3}/96.1) & (96.0/\textbf{96.3}) \\
Capsules & (58.3/76.5) & (62.6/56.5) & (75.0/77.5) & (\textbf{89.3}/\textbf{92.4}) & (81.0/87.5) \\
Macaroni1 & (83.4/93.0) & (81.2/88.3) & (94.6/92.4) & (\textbf{98.5}/\textbf{98.5}) & (97.9/97.6) \\
Macaroni2 & (67.6/73.3) & (67.2/72.5) & (80.8/84.3) & (\textbf{95.0}/\textbf{96.6}) & (82.6/89.5) \\
\bottomrule
\end{tabular}\end{adjustbox}
\end{table*}

In our introduction we mentioned that heuristics can teach us a lot by analyzing the cases where they succeed or fail. There is one particular test case where LWinNN has notably low performance: the multi-instance case, for which we include detailed results in \cref{table:comparison_multi_instance}. Although LWinNN performs well on the `Candle' and `Macaroni1' categories, it has poor results on `Capsules' and `Macaroni2'. For the latter categories, objects are presented unaligned, and transformation invariance would be desired. We see that we outperform PatchCore though, something at odds with the idea that it has stronger transformation invariance than LWinNN. EfficientAD is the only method that performs well here, which might be due to their distillation of pretrained features into specialized features. Though the resulting features are still only translation invariant, they might be specialized to imitate all possible orientations of the objects.

\paragraph{\textbf{Discussion}}
One might think that pre-feature-extraction alignment (Like FYD \cite{zheng2022focus} or GLAD \cite{artola2023glad}) should be able to achieve good results here, but its efficacy for multiple-instance cases is likely to be low. Alignment algorithms typically align the whole image at once, which is incompatible with images of multiple objects that all have to be aligned individually. This has important implications: until the problem of \textit{transformation} invariant feature extraction is solved, we cannot do better than \textit{translation} invariance when using pretrained features. 

It is also important to distinguish between problem statement, benchmark performance, and industry applicability here. Our good performance relies on the fact that the datasets we test on overwhelmingly feature single-instance cases with small translations. Single instance cases with large translations could be solved by alignment-based algorithms, but it is unclear whether this case even exists in practical settings: if a company has the option to capture images of objects one-by-one, would they automatically have the option to align the object in the middle of the frame? Furthermore, does the current ratio of single- to multi-instance images in these benchmarks also reflect the same ratio in industry problems? The answers to these questions (which should come from industry professionals) make the difference between our method being the simplest solution to the problem, or indicative of the datasets not reflecting the real-life setting.
\section{Conclusion}
In this paper we introduced LWinNN, a heuristic that utilizes local window nearest neighbor search. Careful study of related work led to the hypothesis that transformation invariance to only minor translations should lead to good results in many cases. We designed LWinNN for this specific purpose, and demonstrated how it is more accurate and faster than nearest neighbor-based methods that have either full translation invariance or none. This uses the small amount of data much more efficiently and narrows the gap with state-of-the-art methodology, which is much more expensive to train. 

But even though we can boast excellent results, research is more than just benchmark performance. Our assumption of local similarity should fail with spatially varied data, and our results on the multi-instance test cases demonstrate this. The right approach would be to learn which spatial varieties exist in the data instead of assuming it. This hypothetical `right approach' should ideally also be wary of other transformations like rotations and mirrorings, not just translations. Rather than creating \textit{invariance} by storing all orientations separately, it should recognize which patches are semantically identical within the data, thus creating \textit{equivariance} instead. Future research could therefore be directed towards not only learning normality of individual patches, but also combinations of patches, how they connect to each other, their orientation or their location. This could also be a step in the right direction for tackling logical IAD \cite{bergmann2022beyond}. 

The performance of our heuristic is also indicative that new data would be needed to show the validity of this idea: current benchmarks cannot provide a valid proof-of-concept for works claiming to be either in- or equivariant to spatial transformations when minor translation invariance is sufficient. There need to be test cases where minor translation invariance fails and full transformation invariance succeeds to support the claim that these algorithms are doing more than simply minor translation invariance. In conclusion, LWinNN mostly exposes areas of interest for further work, for which our method can hopefully serve as a baseline.
\\\\
\textbf{Acknowledgements }
This work was supported by the KU Leuven Research Fund (C2E/23/007). This research received funding from the Flemish Government under the “Onderzoeksprogramma Artificiële Intelligentie (AI) Vlaanderen” programme.




% LNCS Homepage, \url{http://www.springer.com/lncs}, last accessed 2023/10/25
% \end{thebibliography}

{
    \small
    \bibliographystyle{splncs04}
    \bibliography{references}
}
\end{document}